%% file: main.tex
\documentclass[runningheads]{llncs}
\usepackage{comment}
\usepackage{graphicx}
\usepackage{times}
\usepackage{epsfig}
\usepackage{graphicx}
\usepackage{amsmath}
\usepackage{amssymb}
\usepackage{bbm}
\usepackage{color}
\usepackage{float}
\usepackage{booktabs}
\usepackage{multirow}
\usepackage{hhline}
\usepackage{ltablex}
\usepackage{algorithm,algorithmic}
\usepackage{wrapfig}
\usepackage{graphicx}
\usepackage{subfig}
\usepackage[hidelinks]{hyperref}

\newcommand{\RNum}[1]{\uppercase\expandafter{\romannumeral #1\relax}}

\begin{document}
\title{Cross-Camera Deep Colorization}


\titlerunning{Cross-Camera Deep Colorization}
%
\author{Yaping Zhao\inst{1,2} \and
Haitian Zheng\inst{3} \and
Mengqi Ji\inst{4} \and Ruqi Huang\inst{1,}\thanks{\scriptsize Ruqi Huang is the corresponding author (ruqihuang@sz.tsinghua.edu.cn). This work is supported in part by the Shenzhen Science and Technology Research and Development Funds (JCYJ20180507183706645), in part by the Provincial Key R\&D Program of Zhejiang (Serial No. 2021C01016), and the Shenzhen Key Laboratory of next generation interactive media innovative technolog (Funding No. ZDSYS20210623092001004). The lab website is: \href{http://www.luvision.net}{\textcolor{blue}{http://www.luvision.net}}.}}
\authorrunning{Y. Zhao et al.}
%
\institute{\scriptsize $^1$Tsinghua-Berkeley Shenzhen Institute and Tsinghua Shenzhen International Graduate School,\\\ $^2$Zhejiang Future Technology Institute (Jiaxing),~ $^3$University of Rochester,~ $^4$Beihang University}


\maketitle              
\begin{figure}[!h]
\vspace{-30pt}
    \centering
    \subfloat{\includegraphics[width=0.35\linewidth]{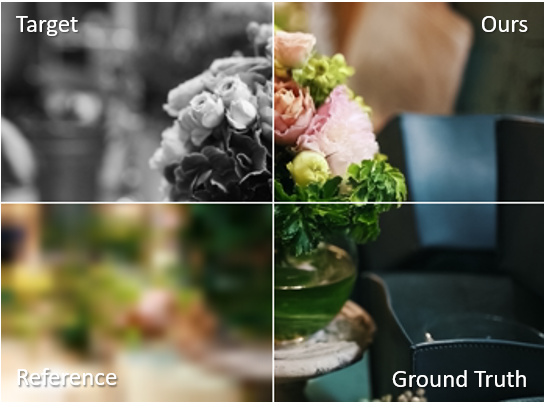}}
    \hfill
  \subfloat{\includegraphics[width=0.65\linewidth]{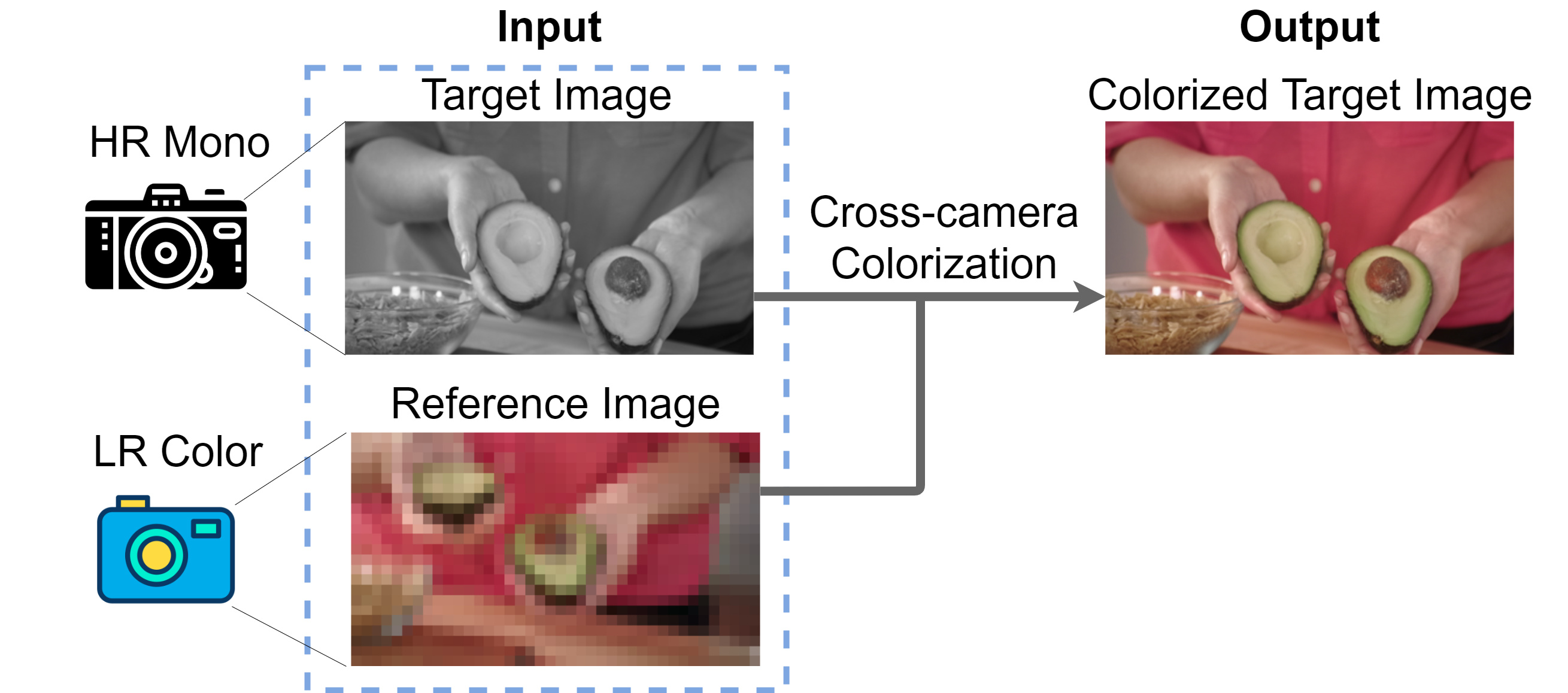}}
  \hfill
  \caption{Aiming at pursuing high-resolution and high color depth simultaneously, we proposed a general binocular imaging framework that performs cross-camera colorization.
    }
\vspace{-30pt}
  \label{fig:teaser}
\end{figure}
\begin{abstract}
    In this paper, we consider the color-plus-mono dual-camera system and propose an end-to-end convolutional neural network to align and fuse images from it in an efficient and cost-effective way. Our method takes cross-domain and cross-scale images as input, and consequently synthesizes HR colorization results to facilitate the trade-off between spatial-temporal resolution and color depth in the single-camera imaging system.
    In contrast to the previous colorization methods, ours can adapt to color and monochrome cameras with distinctive spatial-temporal resolutions, rendering the flexibility and robustness in practical applications. 
    The key ingredient of our method is a cross-camera alignment module that generates multi-scale correspondences for cross-domain image alignment. Through extensive experiments on various datasets and multiple settings, we validate the flexibility and effectiveness of our approach. Remarkably, our method consistently achieves substantial improvements, \emph{i.e.}, around 10dB PSNR gain, upon the state-of-the-art methods. 
    Code is at: \href{https://github.com/IndigoPurple/CCDC}{\textcolor{blue}{github.com/IndigoPurple/CCDC}}. 
    
\vspace{-10pt}
\end{abstract}
\input{section/intro}
\input{section/related}
\input{section/approach}
\input{section/experiment}
\input{section/conclude}
%
%
%
\bibliographystyle{splncs04}
\bibliography{egbib_sim}
%




\end{document}

%% file: section/intro.tex
\vspace{-25pt}
\section{Introduction}
\vspace{-5pt}
Nowadays, it has become a common practice of leveraging fusion of multi-sensory data from camera arrays to improve imaging quality~\cite{tai2008image,cossairt2011gigapixel,brady2012multiscale,ma2014acquisition,cao2011high,jin2020all,wang2020panda,yuan2021modular,li2020zoom,zhao2021cchmt}. In this paper, we consider the color-plus-mono dual-camera system and propose a novel learning-based framework for data fusion. Such a setting enjoys the advantages of photosensibility and high resolution of the monochrome camera, and the color sensibility of the color camera simultaneously. More specifically, our pipeline takes as inputs a high-resolution (HR) grayscale image and a low-resolution (LR) color image, which are taken of the same scene by the respective cameras from similar viewpoints, as shown on the left of Fig.~\ref{fig:teaser}. After fusion, we obtain a HR color image illustrated on the right of Fig.~\ref{fig:teaser}, which to some extent facilitates the trade-off between spatial-temporal resolution and color depth in the single camera due to the space-bandwidth-product (SBP)~\cite{lohmann1996space}. 

In fact, the color-plus-mono dual camera has attracted an increasing amount of attention~\cite{jeon2016stereo,dong2019shoot,liu2016high,mantzel2017shift,sharif2019deep} recently.  
Most of the existing methods focus on addressing the cross-domain image color transfer. For example, traditional methods~\cite{welsh2002transferring,reinhard2001color,ironi2005colorization,tai2005local,gupta2012image,chia2011semantic,liu2008intrinsic,bugeau2013variational} employ global color statistics or low-level feature correspondences while emerging learning-based approaches~\cite{liao2017visual,he2018deep} find high-level feature correspondences between images. 
Nevertheless, they commonly assume similar spatial resolution between data, and therefore can only deal with minor resolution gap, which is typically less than $2\times$. 

On the other hand, it is naturally desirable to retain both high spatial-temporal resolution and high color depth. Thus methods assuming low-resolution gap have to either sacrifice resolution or require high-resolution input from \emph{both} cameras, which can be costly and computationally heavy. Such discrepancy limits their practical applicability, for example, in using cameras with huge-resolution gaps to capture gigapixel video~\cite{yuan2017multiscale,zhang2020multiscale}, or in reducing the budget of a camera system~\cite{wang2016light}.

In contrast, our method enables the imaging system to flexibly employ various cameras with different resolution gaps.
The key ingredient of our method is a cross-camera alignment module that generates multi-scale correspondence for cross-domain image alignment. Without resorting to hand-crafted design on image registration or fusion~\cite{jeon2016stereo,dong2019shoot}, we propose a novel neural network that leverages joint image alignment and fusion for cross-camera colorization.
To improve the correspondence, we design visibility maps computation that explicitly computes and compensates warping errors. Finally, we utilize a warping regularization~\cite{tan2020crossnet++} to further improve the alignment quality.

Extensive experiments are performed to evaluate the proposed method under various settings, i.e., combinations of different resolution gaps, viewpoints, temporal steps, dynamic/static scenes. We test our method on various datasets, including the video dataset Vimeo90k~\cite{xue2019video}, the light field dataset Flower~\cite{srinivasan2017learning}, and the light field video dataset LFVideo~\cite{wang2017light}. Both the quantitative and qualitative experiments show the substantial improvements of our method over the state-of-the-art methods -- remarkably, we achieve around 10dB gain in terms of PSNR in most of the test cases upon the best baseline.

Our main contributions are summarized as follows:
\vspace{-3mm}
\begin{itemize}
    \item A flexible and cost-effective imaging framework that is applicable to various color-plus-mono camera settings with multiple resolution gaps. In particular, our method can adapt to both spatial and temporal resolution gap more than $8\times$.
    \item A novel network design for cross-camera colorization: the cross-camera alignment generates dense correspondence for multi-scale feature alignment; the fusion module compensates alignment error via visibility map computation and performs synthesis; the warping regularization further improves the alignment.
    \item Extensive evaluation on a wide range of settings, \emph{i.e.} different resolution gaps, combinations of viewpoints and temporal steps show the substantial improvements of our method, \emph{i.e.}, around 10dB PSNR gain over the state-of-the-art ones.
\end{itemize}

%% file: section/related.tex
\section{Related Work}
\label{sec:related}
\subsection{Automatic Image Colorization}

Most traditional methods perform colorization by optimization, \textit{e.g.},
Regression Tree Fields (RTFs)~\cite{jancsary2012regression} and graph cuts~\cite{charpiat2008automatic}. 
With deep learning, some approaches~\cite{cheng2015deep,zhang2016colorful} leverage large-scale color image data~\cite{zhou2014learning,russakovsky2015imagenet} to colorize grayscale images automatically.
Yoo \emph{et al.}~\cite{yoo2019coloring} propose a colorization network augmented by external neural memory networks.
Another line of works uses GANs~\cite{goodfellow2014generative,isola2017image,cao2017unsupervised,vitoria2019chromagan} to colorize grayscale images by learning probability distributions.

However, automatic colorization is ill-conditioned since many potential colors can be assigned to the gray pixels of an input image. As a result, these methods tend to generate unnatural colorized images.

\subsection{Reference-based Image Colorization}
According to what is used as a reference, reference-based methods can be divided into strokes/palette-based ones and example-based ones. 

\noindent\textit{Strokes/Palette-based Image Colorization.}
Strokes/palette-based image colorization methods~\cite{levin2004colorization,huang2005adaptive,yatziv2006fast,luan2007natural,chang2015palette} seek to reconstruct color images from the users-provided sparse stroke. 
However, these methods require intensive manual works. More importantly, scribbles and  palettes provide insufficient color information for colorization, leading to unsatisfying results.

\noindent\textit{Example-based Image Colorization.}
To avoid the manual labor of scribbling or palette selection while facilitating controllable colorization, example-based methods are proposed to transfer the color of a reference image to the target image.
The initial works~\cite{reinhard2001color,welsh2002transferring} attempt to match color statistics globally.
Targeting at more accurate color transfer, later works~\cite{ironi2005colorization,tai2005local,gupta2012image,chia2011semantic,liu2008intrinsic,bugeau2013variational} utilize hand-crafted feature correspondences, \textit{e.g.}, SIFT, Gabor wavelet to enforce local color consistency.
Recently, He \emph{et al.}~\cite{he2017neural} and Liao \emph{et al.}~\cite{liao2017visual} perform dense patch matching for color transfer.
However, the patch-based correspondence~\cite{he2017neural,liao2017visual} are inherently inefficient to compute. Furthermore,
the inter-patch misalignment may hurt the image synthesis. 
Xiao \emph{et al.}~\cite{xiao2020example} propose a self-supervised approach that reconstructs a color image from its grayscale version and global color distribution coding. Although obtaining impressive results, the proposed approach does not utilize the spatial correspondence between two images for local colorization.

\subsection{Flow-based or non-rigid correspondences}
The cross-camera image colorization is also related to flow-based and non-rigid correspondence estimation. Specifically, back-warping reference images using estimated optical flow fields~\cite{liu2010sift,ilg2017flownet,zhao2021efenet,zhao2022manet} can serve to align images from two viewpoints. In addition, HaCohen \emph{et al.}~\cite{hacohen2011non} showcase the usage of estimating non-rigid dense correspondence for image enhancement. However, due to the domain and resolution gap between the two cameras, the above approaches fail in finding correct matching and result in poor performance under our cross-domain and cross-resolution setting. 

%% file: section/approach.tex
\section{Approach}
\label{sec:approach}
\vspace{-2mm}
Assume two images from different cameras capturing the same scene at similar viewpoints are given -- an LR color image as the reference and an HR gray image as the target. 
We denote the target single-channel image as $I_1 \in {\mathbb{R}}^{sH \times sW \times 1}$ and the three-channel reference as $I_2 \in {\mathbb{R}}^{H \times W \times 3}$, where $s\geq 1$ is the scale factor representing the spatial-resolution gap in the dual camera, $H$ and $W$ are the horizontal and vertical spatial-resolution, respectively. Similarly, the ground-truth color image is denoted as $I_g \in {\mathbb{R}}^{sH \times sW \times 3}$, where $I_g$ shares the same viewpoint with the target image $I_1$. 
Our goal is to generate an HR color image $I_c \in {\mathbb{R}}^{sH \times sW \times 3}$ as similar as possible to $I_g$.

\begin{figure*}[t]
	\centering
  \includegraphics[width=\linewidth]{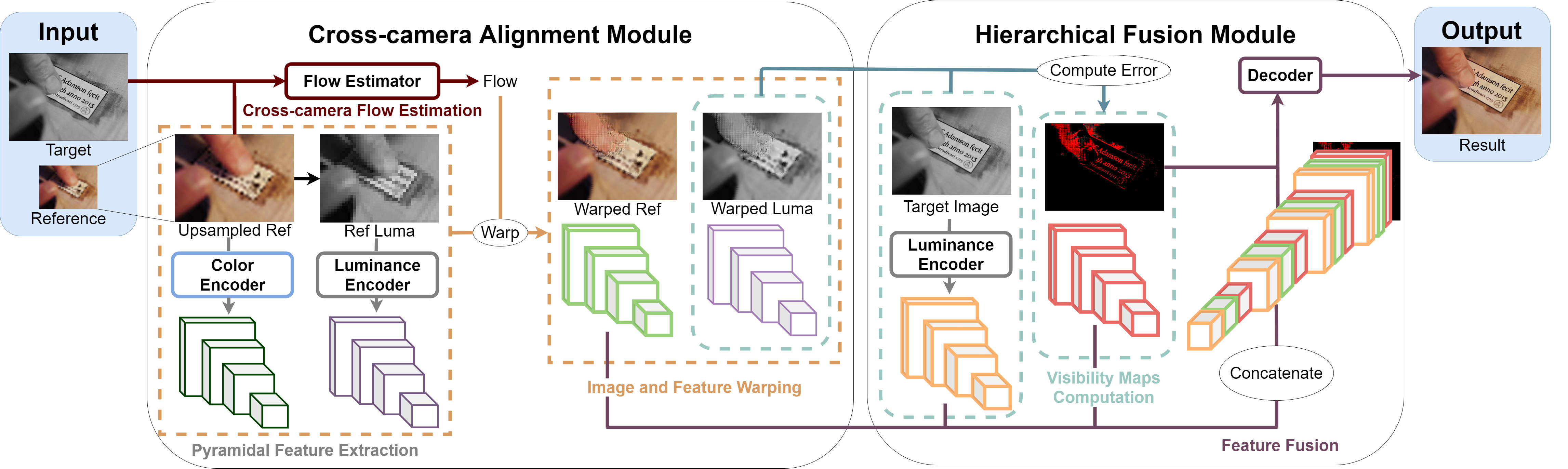}
  \caption{Our network contains two modules: 1) the alignment module, which performs non-rigid transformation on the cross-camera image inputs and extracted features; 2) the fusion module, which performs features fusion and colorization synthesis. }
  \vspace{-5mm}
  \label{fig:pipeline}
\end{figure*}

To achieve this goal, we propose an end-to-end and fully convolutional deep neural network (see illustration in Fig.~\ref{fig:pipeline}). Our network contains a cross-camera alignment module and a hierarchical fusion module: the former (Sec.~\ref{sec:alignment}) consists of a color encoder ${Net}_E^c$, a luminance encoder ${Net}_E^l$, and a flow estimator $Net_{flow}$; the latter (Sec.~\ref{sec:fusion}) consists of a decoder $Net_D$. 
On top of that, our network trains jointly the warping flow estimator and colorization neural networks by combining warping loss $\mathcal{L}_w$ and colorization loss $\mathcal{L}_c$ (Sec.~\ref{sec:loss}).

\vspace{-2mm}
\subsection{Cross-camera Alignment Module}
\label{sec:alignment}

This module is designed to perform the temporal and spatial alignment. 
As shown on the left of Fig.  \ref{fig:pipeline}, we first extract feature maps of input images and then utilize a flow estimator to generate the cross-camera correspondence at multiple scales. After flow estimation, we use cross-camera warping to perform the non-rigid transformation. In the following, we provide details of each part. 

\vspace{-2mm}
\paragraph{Pyramidal Feature Extraction.} 
Considering the different resolutions of the input images, we first upsample the reference $I_2$ to the same resolution as $I_1$ via bicubic upsampling,
denoted by $I_2^\uparrow$. $I_1$ and $I_2^{\uparrow}$ are of the same resolution but belong to different image modalities,
thus we design two encoders to extract their pyramidal features respectively:

\begin{align}
\begin{aligned}
&\{F_i^l\} = {Net}_E^l(I_1),
&\{F_i^c\} = {Net}_E^c(I_2^\uparrow),\ \ \ i = 1, 2, 3, 4,
\end{aligned}
\end{align}
where $F_i^l$ (resp. $F_i^c$) is the feature map of target image $I_1$ (resp. $I_2$) at scale $i$.
 
To measure the image alignment across different modalities, we convert the upsampled reference image $I_2^\uparrow$ from RGB color space to YUV~\cite{union1992encoding} color space.
Thus, $I_2^\uparrow$ can be separated into a luminance component and two chrominance components. Discarding chrominance components, we retain the luminance components denoted as $I^Y_2$, which is a grayscale counterpart of the upsampled reference image $I_2^\uparrow$.

Then the luminance encoder is also used to extract multi-scale feature maps of $I^Y_2$, which is utilized latter in Sec. \ref{sec:fusion} to calculate warping errors on feature domain:
\begin{align}
\begin{aligned}
\{F_i^Y\} = {Net}_E^l(I^Y_2),\ \ \ i = 1, 2, 3, 4.
\end{aligned}
\end{align}

\paragraph{Cross-camera Flow Estimation.} For image alignment, we adopt the widely used FlowNetS~\cite{dosovitskiy2015flownet} as our flow estimator, ${Net}_{flow}$, to generate the dense cross-camera correspondence at multiple scales. Tailored for our setting, we change the input channel number of the first convolutional layer of FlowNetS from $6$ to $4$, and obtain the following flow fields:

\begin{align}
\begin{aligned}
\label{eq:flow}
\{f_i\} =  {{Net}_{flow}}(I_1, I_2^\uparrow),\ \ \ i = 0, 1, 2, 3, 4,
\end{aligned}
\end{align}
where $f_i$ is the estimated flow field at scale $i$.

\paragraph{Image and Features Warping.} To perform the temporal and spatial alignment with the estimated flow fields, we utilize a warping operation similar to~\cite{zheng2018crossnet}. 
More specifically, our warping operation considers the cross-camera flow field $f$:

\begin{align}
\begin{aligned}
\label{eq:warp}
\widetilde{I} = \mathcal{W}(I, f), 
\end{aligned}
\end{align}
where $\mathcal{W}(I, f)$ denotes the result of warping the input $I$ using the flow field $f$.

After flow estimation, we perform the warping operation on the reference image features $F_i^c$ and corresponding luminance features $F_i^Y$. Using the multi-scale flow $f_i$ in Eq. \ref{eq:flow}, we generate the
temporally and spatially aligned features $\{\widetilde{F}_i^c\}$ and $\{\widetilde{F}_i^Y\}$:
\begin{align}
\begin{aligned}
\label{eq:align}
&\widetilde{F}_i^c =  \mathcal{W}(F_i^c, f_i),
&\widetilde{F}_i^Y =  \mathcal{W}(F_i^Y, f_i),\ \ \ i = 1, 2, 3, 4.
\end{aligned}
\end{align}

To measure image alignment latter in Sec. \ref{sec:fusion}, we also perform the warping operation on the image domain. According to Eq. \ref{eq:warp}, we have:
\begin{align}
\begin{aligned}
\label{eq:warpi}
\widetilde{I}^Y_2 = \mathcal{W}(I^Y_2, f_0),
\end{aligned}
\end{align}
where $f_0$ is the estimated flow field at scale $0$ in Eq. \ref{eq:flow}, $\widetilde{I}^Y_2$ is the warping result utilized latter in Sec. \ref{sec:fusion} to calculate warping errors on the image domain.

\subsection{Hierarchical Fusion Module}
\label{sec:fusion}

\paragraph{Visibility Maps Computation.} 

On the image domain, the warping error indicates the different light intensity between the target and the reference, \emph{i.e.}, optical visibility. While on the feature domain, warping error represents the different activation value of the feature maps, \emph{i.e.}, feature recognition. 
We combine the errors from both perspectives and define the multi-scale visibility maps $\{V_i\}$ as warping errors on both domains:

\begin{align}
\begin{aligned}
&V_0 = \widetilde{I}^Y_2 - I_1,
&V_i = \widetilde{F}_i^Y - {F_i^l},\ \ \ i = 1, 2, 3, 4, 
\end{aligned}
\end{align}
where $V_i$ is the warping error at scale $i$.

\begin{wrapfigure}{r}{0.5\textwidth}
\vspace{-12mm}
  \begin{center}
    \includegraphics[width=0.5\textwidth]{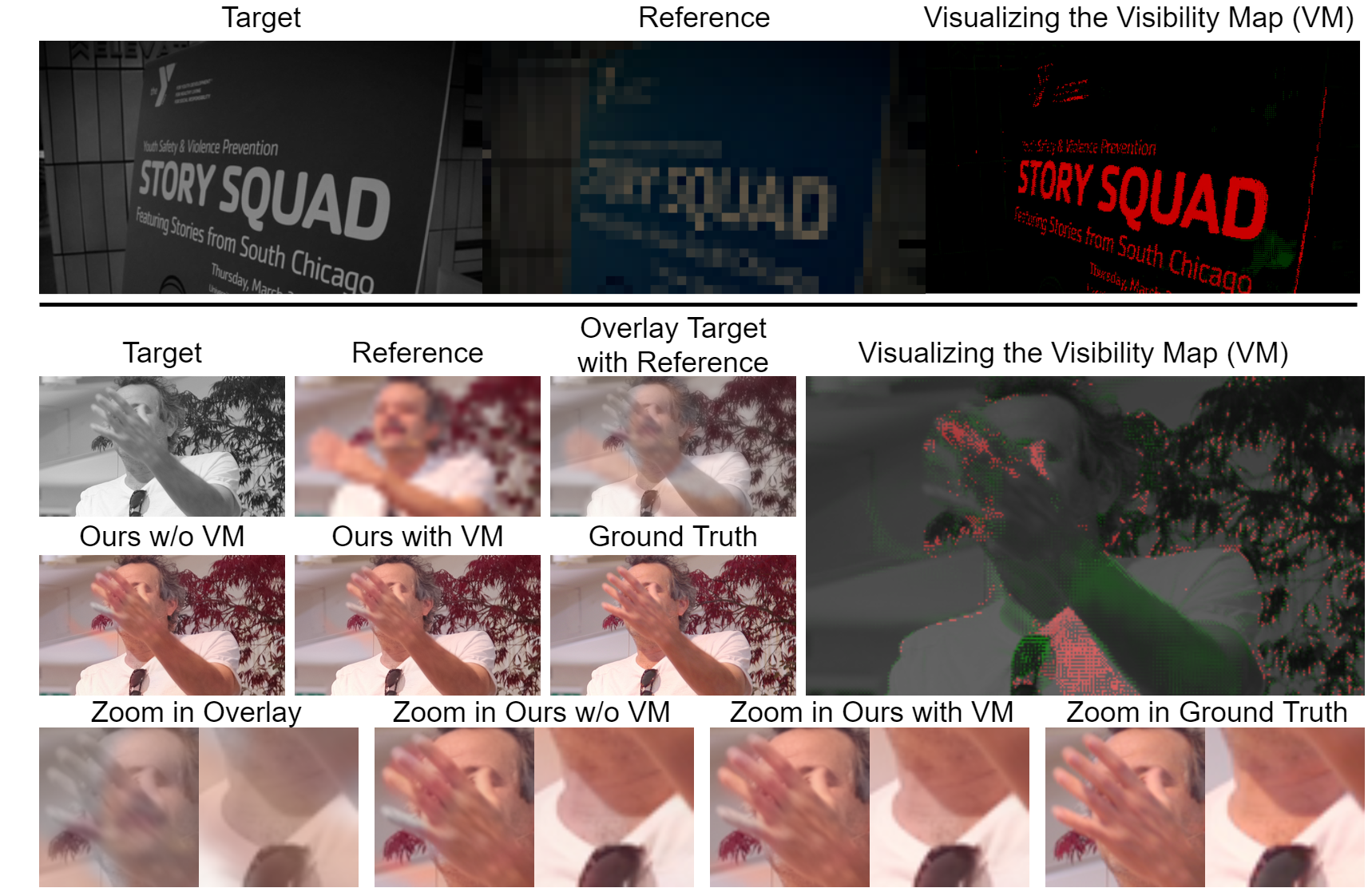}
  \end{center}
  \vspace{-5mm}
  \caption{Visualizing the visibility map on image domain. Red: invisible regions in the reference image; green: invisible regions in the target image. Zoom in to see details.}
  \vspace{-10mm}
    \label{fig:visibility}
\end{wrapfigure}


To give an intuition of the visibility maps, we visualize it on the image domain in Fig. \ref{fig:visibility}. 
There are lost details caused by the low-quality reference (\emph{e.g.}, blurry words shown on the top of Fig.~\ref{fig:visibility}), motion blur caused by fast-moving objects (\emph{e.g.}, waving hand shown on the bottom of Fig.~\ref{fig:visibility}), local occlusion caused by motion or parallax (\emph{e.g.}, the garment occluded by the hand shown on the bottom of Fig.~\ref{fig:visibility}). 
The pixel-wise positive and negative values of visibility maps represent the invisible regions in the reference and the target image, respectively.

\paragraph{Feature Fusion.}
In the end, we design a U-Net~\cite{ronneberger2015u} like decoder to fuse feature maps and visibility maps, and synthesize the colorization result. As shown on the right of Fig. \ref{fig:pipeline}, the target features $\{F_i^l\}$, warped reference features $\{\widetilde{F}_i^c\}$ and visibility maps $\{V_i\}$ are concatenated as the input of fusion decoder. Finally we obtain the result $I_c$ by:
\begin{align}
\begin{aligned}
\label{eq:final}
I_c = {Net}_D(\{F_i^l\}, \{\widetilde{F}_i^c\}, \{V_j\}),\ \ \ i=1,2,3,4, j=0,1,2,3,4.
\end{aligned}
\end{align}

\subsection{Loss Function}\label{sec:loss}
We use two loss functions: warping loss and colorization loss. The former encourages the flow estimator to generate precise cross-camera correspondence for image alignment. The latter is responsible for the final synthesized image.

\paragraph{Warping Loss.} 
Since the ground-truth flow is unavailable, it is difficult to train the flow estimator in an unsupervised fashion. 
To solve this problem, 
we adopt the warping loss from~\cite{tan2020crossnet++}.
Specifically, since the input images capture the same scene, it is reasonable to require the warped-upsampled reference image owning a intensity distribution similar to the ground truth ${I_g}$ as much as possible. Thus our warping loss is defined as:

\begin{align}
\begin{aligned}
&\mathcal{L}_w =  \frac{1}{2N}\sum_{i=1}^N\sum_j {|| {\widetilde{I}^{\uparrow}_2}{}^{(i)} - {I}_g^{(i)}||}_2^2,
&\widetilde{I}^\uparrow_2 = \mathcal{W}(I_2^\uparrow, f_0),
\end{aligned}
\end{align}
where $N$ is the sample number, $i$ iterates over training samples, $f_0$ is the estimated flow field at scale $0$ in Eq. \ref{eq:flow}.


\paragraph{Colorization Loss.} Given the network prediction $\widetilde{I_c}$ and the ground truth $I_c$, the colorization loss is defined as:
\begin{align}
\begin{aligned}
\mathcal{L}_c =  \frac{1}{N}\sum_{i=1}^N\sum_j\rho( I_c^{(i)} - {I}_g^{(i)}),
\end{aligned}
\end{align}
where $\rho(x)=\sqrt{x^2 + {0.001}^2}$ is the Charbonnier penalty function~\cite{bruhn2005lucas}, $\widetilde{I}_c$ is obtained from Eq.~\ref{eq:final}, $N$ is the sample number, $i$ iterates over training samples.

%% file: section/experiment.tex
\begin{figure*}[]
\vspace{-5mm}
  \centering
  \includegraphics[width=\linewidth]{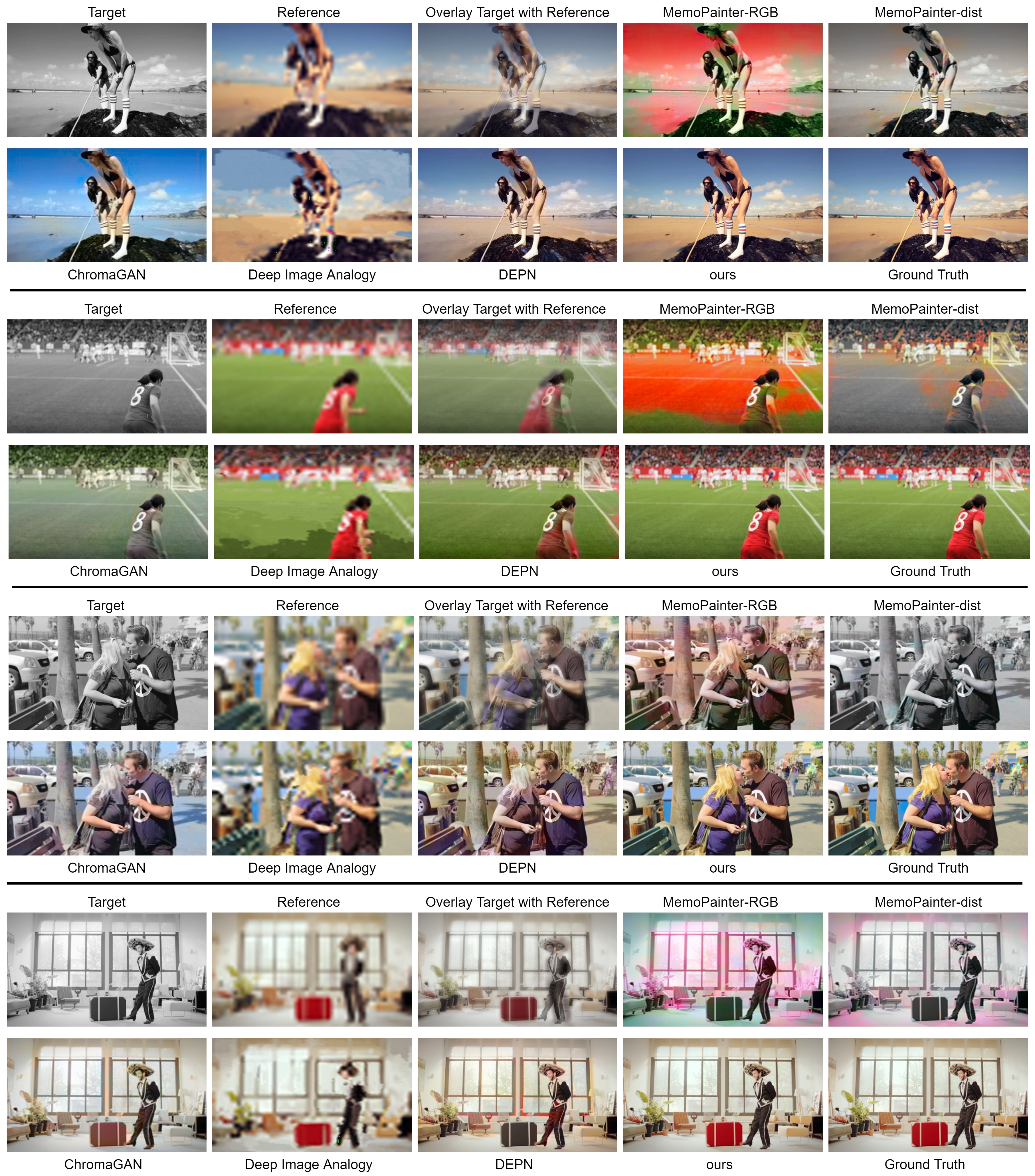}
  \caption{Colorization comparisons on Vimeo90K dataset under cross-scale $8\times$ settings. 
  }
  \label{fig:vimeo}
\end{figure*}

\begin{figure*}[htbp!]
\vspace{-3mm}
  \centering
  \includegraphics[width=\linewidth]{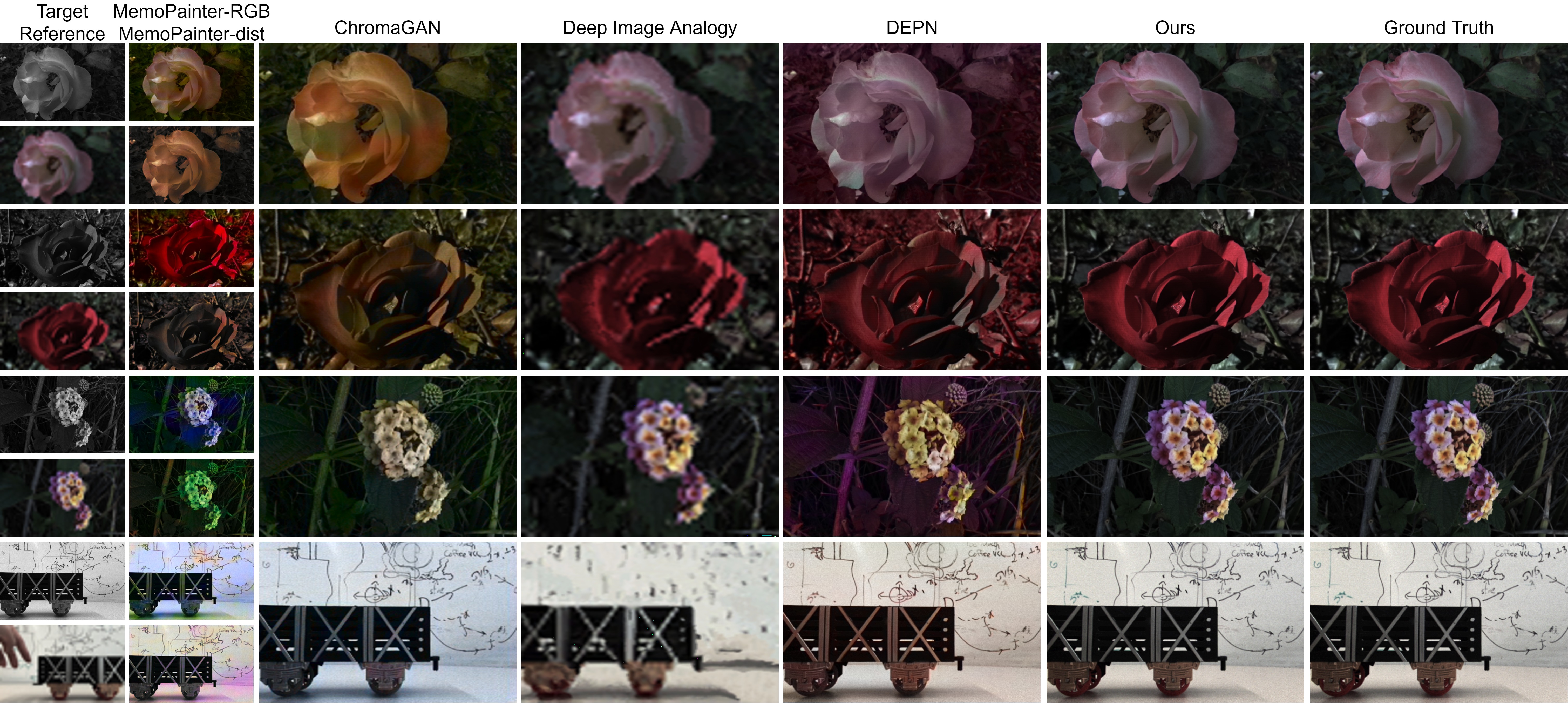}
  \caption{Comparison under cross-scale $8\times$ settings on Flower (the 1st, 2nd, 3rd rows) and LFVideo (the last row) dataset, respectively. 
  }
  \vspace{-3mm}
  \label{fig:flower}
\end{figure*}


\section{Experiments}
\label{sec:experiment}
\vspace{-2mm}
\subsection{Dataset}
\vspace{-2mm}
\paragraph{Video Dataset.} 
The Vimeo90K~\cite{xue2019video} dataset consists of videos that cover a large variety of scenes and actions. Following~\cite{xue2019video}, we selected $66,178$ video sequences, each of which contains $7$ frames with resolution of $448\times256$. To construct the training and testing datasets, we randomly divide it into $60,000$ sequences for training and $6,178$ sequences for testing. For training, we downsample the first frame at each video sequence as a reference image and randomly select the $t$th$(1<t<8)$ frame converted to the grayscale image as a target. For testing, reference images are sampled from the first video frame, while the target images are from the second and the last frame. See the supplementary material for training details.


\vspace{-2mm}
\paragraph{Light-field Dataset.} The Flower dataset~\cite{srinivasan2017learning} contains flowers and plants light-field images with the $376 \times 541$ spatial resolution, and $14 \times 14$ angular samples. Following~\cite{srinivasan2017learning}, we extract the central $8 \times 8$ grid of angular sample, and randomly selected 343 samples for evaluations. The images at viewpoints $(1,1)$ and $(7,7)$ are converted to grayscale as target, and images at viewpoint $(0, 0)$ are downsampled as reference.

\vspace{-2mm}
\paragraph{Light-field Video Dataset.} The LFVideo dataset~\cite{wang2017light} contains real-scene light-field videos with the spatial resolution as $376\times541$, while the angular samples are $8\times8$. For evaluations, we randomly selected $270$ video frames. The images of the $t$th$(t=2, 9)$ frame at viewpoints $(i, i), i=1,7$  are converted to monochrome as target, and images of the first frame at viewpoint $(0, 0)$ are downsampled as reference.

\vspace{-2mm}
\subsection{Comparison to State-of-the-Art Methods}
\vspace{-2mm}
\begin{table}[htbp!]
\begin{minipage}{\linewidth}
\vspace{-5mm}
\caption{Quantitative evaluations of the state-of-the-art automatic and example-based algorithms on different datasets, in terms of NRMSE/PSNR/SSIM/LPIPS for different scale factors, frame gaps and parallax settings respectively.}
 \label{table:evaluation}
\centering
\tabcolsep=0.35cm
\resizebox{\textwidth}{!}{ 
\begin{tabular}{cccccccccc}
\hline
\multirow{2}{*}{Dataset} & \multicolumn{2}{c}{Target Position} & \multirow{2}{*}{Scale} & \multirow{2}{*}{Methods}                                     & \multicolumn{5}{c}{Quantitative evaluations}                                                                                                                                    \\
\cmidrule(l{4pt}r{4pt}){2-3} \cmidrule(l{4pt}r{4pt}){6-10}& Frame            & View            &                        &                                                              & NRMSE                      & PSNR                       & SSIM                       & LPIPS                                 & Runtime                    \\ \hline
\multirow{18}{*}{Vimeo}   & 2              & -                & N/A                      & MemoPainter-RGB~\cite{yoo2019coloring}  & 0.3463                   & 19.9920                    & 0.7989                     & 0.4206                     & 0.5124                   \\
                          & 2              & -                & N/A                      & MemoPainter-dist~\cite{yoo2019coloring}                       & 0.2352                   & 22.5768                    & 0.8751                     & 0.3015                     & 0.5620                   \\
                          & 2              & -                & N/A                      & ChromaGAN ~\cite{vitoria2019chromagan}                        & 0.2097                   & 23.5276                    & 0.8697                     & 0.2917                     & 2.6915                   \\
                          & 2              & -                & $4\times$              & Deep Image Analogy~\cite{liao2017visual}                      & 0.1544                   & 25.6775                    & 0.7741                     &  0.3184                   & 219.5779                 \\
                          & 2              & -                & $4\times$              & DEPN ~\cite{xiao2020example}                                  & 0.1313                   & 27.5916                    & 0.9313                     & 0.1585                     & 0.3425                   \\
                          & 2              & -                & $4\times$              & Ours                                                         & \textbf{0.0227}  & \textbf{43.2263}   & \textbf{0.9884}    & \textbf{0.0157}   & \textbf{0.0838}  \\
                          & 2              & -                & $8\times$              & DEPN ~\cite{xiao2020example}                                  & 0.1316                   & 27.5724                    & 0.9310                     & 0.1588                     & 0.3525                   \\
                          & 2              & -                & $8\times$              & Deep Image Analogy~\cite{liao2017visual}                      &  0.1951                  &  23.5943                  & 0.6954                    & 0.4142                    & 203.4619                 \\
                          & 2              & -                & $8\times$              & Ours                                                         & \textbf{0.0275} & \textbf{41.6039}  &  \textbf{0.9845} & \textbf{0.0241}  & \textbf{0.0847} \\
                          \cmidrule(l{4pt}r{4pt}){2-10}
                          & 7              & -                & N/A                      & MemoPainter-RGB ~\cite{yoo2019coloring}                       &  0.3150                  & 20.0014                    & 0.7995                     & 0.4201                     & 0.5301                   \\
                          & 7              & -                & N/A                      & MemoPainter-dist~\cite{yoo2019coloring}                       &  0.2326                  & 22.4321                    & 0.8738                     & 0.3028                     & 0.5483                   \\
                          & 7              & -                & N/A                      & ChromaGAN ~\cite{vitoria2019chromagan}                        & 0.2064                   &  23.5428                   & 0.8696                     & 0.2909                    & 2.7384                   \\
                          & 7              & -                & $4\times$              & Deep Image Analogy~\cite{liao2017visual}                     & 0.2063                   & 23.0956                    & 0.7227                    & 0.3529                    & 214.3759                 \\
                          & 7              & -                & $4\times$              & DEPN ~\cite{xiao2020example}                                  & 0.1306                   & 27.5174                    & 0.9313                     & 0.1593                     & 0.3548                   \\
                          & 7              & -                & $4\times$              & Ours                                                         & \textbf{0.0380}  & \textbf{39.5223}   & \textbf{0.9823}   & \textbf{0.0321}   & \textbf{0.0846}  \\
                          & 7              & -                & $8\times$              & Deep Image Analogy~\cite{liao2017visual}                     &  0.2356                  &  21.8671                  &  0.66771                   & 0.4260                    & 220.8749                 \\
                          & 7              & -                & $8\times$              & DEPN~\cite{xiao2020example}                                  & 0.1309                   & 27.4989                    & 0.9311                     & 0.1596                     & 0.3478                   \\
                          & 7              & -                & $8\times$              & Ours                                                         & \textbf{0.0408} & \textbf{38.6736}  &  \textbf{0.9796} & \textbf{0.0382}  & \textbf{0.0891} \\ \hline
\multirow{18}{*}{Flower}  & -                & (1,1)            & N/A                      & MemoPainter-RGB~\cite{yoo2019coloring}                                      & 0.4172                   & 22.0304                    & 0.7508                    & 0.3389                    & 0.5623                   \\
                          & -                & (1,1)            & N/A                      & MemoPainter-dist~\cite{yoo2019coloring}                                           & 0.3237                   & 24.2114                    & 0.8822                     & 0.2668                    & 0.6184                   \\
                          & -                & (1,1)            & N/A                      & ChromaGAN~\cite{vitoria2019chromagan}                                        & 0.3046                   & 24.4863                    &  0.8407                    & 0.3081                     & 1.1562                   \\
                          & -                & (1,1)            & $4\times$              & Deep Image Analogy~\cite{liao2017visual}                                 & 0.1874                   & 28.6252                    & 0.8065                     & 0.2581                    & 411.6000                 \\
                          & -                & (1,1)            & $4\times$              & DEPN~\cite{xiao2020example}                                & 0.1797                   & 29.0692                    & 0.9286                     & 0.1711                     & 0.4899                   \\
                          & -                & (1,1)            & $4\times$              & Ours                                                         & \textbf{0.0205}  & \textbf{45.9354}   & \textbf{0.9938}    & \textbf{0.0041}   & \textbf{0.1173} \\
                          & -                & (1,1)            & $8\times$              & Deep Image Analogy~\cite{liao2017visual}                                 &   0.2620                 &  25.6807                 & 0.6976                     &  0.4029                   & 404.4650                 \\
                          & -                & (1,1)            & $8\times$              & DEPN~\cite{xiao2020example}                                & 0.1797                   & 29.0730                    & 0.9285                     & 0.1710                     & 0.4729                   \\
                          & -                & (1,1)            & $8\times$              & Ours                                                         & \textbf{0.0255} & \textbf{43.8623}  &  \textbf{0.9923} & \textbf{0.0077}  & \textbf{0.1171}  \\
                          \cmidrule(l{4pt}r{4pt}){2-10}
                          & -                & (7,7)            & N/A                      & MemoPainter-RGB~\cite{yoo2019coloring}                                      &  0.4216                  & 21.9233                    & 0.7525                    &  0.3424                   & 0.5912                   \\
                          & -                & (7,7)            & N/A                      & MemoPainter-dist~\cite{yoo2019coloring}                                           & 0.3158                   & 24.4801                    & 0.8858                     & 0.2617                    & 0.6034                     \\
                          & -                & (7,7)            & N/A                      & ChromaGAN~\cite{vitoria2019chromagan}                                        & 0.3065                   & 24.4616                   &  0.8404                    & 0.3081                     & 1.5492                   \\
                          & -                & (7,7)            & $4\times$              & Deep Image Analogy~\cite{liao2017visual}                                 & 0.2532                   & 25.9906                    & 0.7100                     & 0.2954                    & 409.2839                 \\
                          & -                & (7,7)            & $4\times$              & DEPN~\cite{xiao2020example}                                & 0.1775                   & 29.2156                    & 0.9300                     & 0.1663                     & 0.4725                   \\
                          & -                & (7,7)            & $4\times$              & Ours                                                         & \textbf{0.0237}  &  \textbf{44.8138}  & \textbf{0.9931}    & \textbf{0.0060}   & \textbf{0.1170}  \\
                          & -                & (7,7)            & $8\times$              & Deep Image Analogy~\cite{liao2017visual}                                 &   0.2903                 &  24.7787                 & 0.6703                     & 0.4006                    & 405.3829                 \\
                          & -                & (7,7)            & $8\times$              & DEPN~\cite{xiao2020example}                                & 0.1774                   & 29.2186                    & 0.9299                     & 0.1663                     & 0.4793                     \\
                          & -                & (7,7)            & $8\times$              & Ours                                                         & \textbf{0.0276} & \textbf{43.2567} &  \textbf{0.9920} & \textbf{0.0092}  & \textbf{0.1175} \\ \hline
\multirow{36}{*}{LFVideo} & 2              & (1,1)            & N/A                      & MemoPainter-RGB~\cite{yoo2019coloring}                                      & 0.4190                   & 21.4680                    & 0.7227                    & 0.3470                     & 0.6633                   \\
                          & 2              & (1,1)            & N/A                      & MemoPainter-dist~\cite{yoo2019coloring}                                           & 0.3317                   & 23.7711                    & 0.8565                     & 0.2181                     & 0.6927                   \\
                          & 2              & (1,1)            & N/A                      & ChromaGAN~\cite{vitoria2019chromagan}                                        & 0.1982                   & 27.6557                    &  0.8592                   & 0.2120                     & 0.5773                   \\
                          & 2              & (1,1)            & $4\times$              & Deep Image Analogy~\cite{liao2017visual}                                 & 0.3304                   & 23.2553                    & 0.6374                     & 0.4040                    & 755.4143                 \\
                          & 2              & (1,1)            & $4\times$              & DEPN~\cite{xiao2020example}                                & 0.1129                   & 32.7250                    & 0.9668                     & 0.0933                     & 0.4824                   \\
                          & 2              & (1,1)            & $4\times$              & Ours                                                         & \textbf{0.0459}  & \textbf{41.0111}   & \textbf{0.9857}    & \textbf{0.0288}   & \textbf{0.1173} \\
                          & 2              & (1,1)            & $8\times$              & Deep Image Analogy~\cite{liao2017visual}                                 &   0.3695                 &   22.3074                & 0.5970                     & 0.4944                    & 748.9276                 \\
                          & 2              & (1,1)            & $8\times$              & DEPN~\cite{xiao2020example}                                & 0.1129                   & 32.7220                    & 0.9669                     & 0.0933                     & 0.4793                   \\
                          & 2              & (1,1)            & $8\times$              & Ours                                                         & \textbf{0.0492} & \textbf{40.3612} & \textbf{0.9849}  &  \textbf{0.0312} & \textbf{0.1167}  \\
                          \cmidrule(l{4pt}r{4pt}){2-10}
                          & 2              & (7,7)            & N/A                      & MemoPainter-RGB~\cite{yoo2019coloring}                                      &  0.4261                  & 21.3552                    & 0.7125                    & 0.3527                     & 0.6325                   \\
                          & 2              & (7,7)            & N/A                      & MemoPainter-dist~\cite{yoo2019coloring}                                           & 0.3058                   & 24.4653                    & 0.8662                     &  0.2037                    & 0.6429                   \\
                          & 2              & (7,7)            & N/A                      & ChromaGAN~\cite{vitoria2019chromagan}                                        & 0.1959                   & 27.8429                   &  0.8609                    & 0.2097                     & 0.5826                   \\
                          & 2              & (7,7)            & $4\times$              & Deep Image Analogy~\cite{liao2017visual}                                 & 0.3394                   & 22.9054                    & 0.6315                    & 0.4041                    & 751.4360                 \\
                          & 2              & (7,7)            & $4\times$              & DEPN~\cite{xiao2020example}                                & 0.1104                   & 32.9745                    & 0.9679                     & 0.0880                     & 0.4529                   \\
                          & 2              & (7,7)            & $4\times$              & Ours                                                         & \textbf{0.0448}  &  \textbf{41.0335}  & \textbf{0.9859}    & \textbf{0.0274}   & \textbf{0.1185} \\
                          & 2              & (7,7)            & $8\times$              & Deep Image Analogy~\cite{liao2017visual}                                 &   0.3718                 &  22.0710                 & 0.5977                     & 0.4920                    & 753.5917                 \\
                          & 2              & (7,7)            & $8\times$              & DEPN~\cite{xiao2020example}                                & 0.11070                  & 32.9560                    & 0.9680                     & 0.0883                     & 0.4826                   \\
                          & 2              & (7,7)            & $8\times$              & Ours                                                         & \textbf{0.0480} & \textbf{40.4257} &  \textbf{0.9853} &  \textbf{0.0301} & \textbf{0.1179}  \\
                          \cmidrule(l{4pt}r{4pt}){2-10}
                          & 9              & (1,1)            & N/A                      & MemoPainter-RGB~\cite{yoo2019coloring}                                      &  0.4501                  & 21.1598                    & 0.6991                   & 0.3608                     & 0.6530                   \\
                          & 9              & (1,1)            & N/A                      & MemoPainter-dist~\cite{yoo2019coloring}                                           &  0.3290                  & 23.7217                    & 0.8582                     & 0.2204                     & 0.6498                   \\
                          & 9              & (1,1)            & N/A                      & ChromaGAN~\cite{vitoria2019chromagan}                                        & 0.2117                   & 27.5192                    &   0.8512                  & 0.2244                     & 0.5728                   \\
                          & 9              & (1,1)            & $4\times$              & Deep Image Analogy~\cite{liao2017visual}                                 & 0.3270                   & 23.5479                    & 0.6404                     & 0.4102                    & 749.7418                 \\
                          & 9              & (1,1)            & $4\times$              & DEPN~\cite{xiao2020example}                                & 0.1165                   & 32.5500                    & 0.9670                     & 0.0917                     & 0.4692                   \\
                          & 9              & (1,1)            & $4\times$              & Ours                                                         & \textbf{0.0508}  & \textbf{40.7632}  & \textbf{0.9853}    &  \textbf{0.0323}  & \textbf{0.1169} \\
                          & 9              & (1,1)            & $8\times$              & Deep Image Analogy~\cite{liao2017visual}                                 & 0.3709                   &  22.4716                & 0.5979                     &  0.4954                   & 747.7591                 \\
                          & 9              & (1,1)            & $8\times$              & DEPN~\cite{xiao2020example}                                & 0.1166                   & 32.5409                    & 0.9671                     & 0.0919                     & 0.4792                   \\
                          & 9              & (1,1)            & $8\times$              & Ours                                                         & \textbf{0.0538} & \textbf{40.1106} & \textbf{0.9845}  &  \textbf{0.0349} & \textbf{0.1167}  \\
                          \cmidrule(l{4pt}r{4pt}){2-10}
                          & 9              & (7,7)            & N/A                      & MemoPainter-RGB~\cite{yoo2019coloring}                                      &  0.4104                  & 21.9786                   & 0.7211                     & 0.3464                     & 0.6009                   \\
                          & 9              & (7,7)            & N/A                      & MemoPainter-dist~\cite{yoo2019coloring}                                           & 0.3271                   & 23.9009                    & 0.8628                     & 0.2170                    & 0.6947                   \\
                          & 9              & (7,7)            & N/A                      & ChromaGAN~\cite{vitoria2019chromagan}                                        & 0.2112                   & 27.5598                   &  0.8542                    & 0.2248                     & 0.5927                   \\
                          & 9              & (7,7)            & $4\times$              & Deep Image Analogy~\cite{liao2017visual}                                 & 0.3221                   & 23.7166                    & 0.6447                     & 0.4058                    & 747.9275                 \\
                          & 9              & (7,7)            & $4\times$              & DEPN~\cite{xiao2020example}                                & 0.1130                   & 32.7364                    & 0.9676                     & 0.0882                     & 0.4865                   \\
                          & 9              & (7,7)            & $4\times$              & Ours                                                         & \textbf{0.0492}  & \textbf{40.8788}  & \textbf{0.9858}    & \textbf{0.0304}   & \textbf{0.1420}  \\
                          & 9              & (7,7)            & $8\times$              & Deep Image Analogy~\cite{liao2017visual}                                 &  0.3564                  &  22.8959                 & 0.6035                     & 0.4921                    & 750.8465                 \\
                          & 9              & (7,7)            & $8\times$              & DEPN~\cite{xiao2020example}                                & 0.1132                   & 32.7194                    & 0.9678                     & 0.0881                     & 0.4539                   \\
                          & 9              & (7,7)            & $8\times$              & Ours                                                         & \textbf{0.0518} & \textbf{40.3270} &  \textbf{0.9851} & \textbf{0.0328}  & \textbf{0.1453} \\ \bottomrule
\end{tabular}
}
\end{minipage}
\vspace{-0.2cm}
\end{table}


Our method is compared against the state-of-the-art example-based colorization methods, namely Deep Image Analogy~\cite{liao2017visual} and DEPN~\cite{xiao2020example}, and the recent automatic colorization approaches, including MemoPainter~\cite{yoo2019coloring} and ChromaGAN~\cite{vitoria2019chromagan}. Following~\cite{yoo2019coloring}, we evaluated MemoPainter using RGB information and color distribution for color features, respectively.
Both the quantitative evaluation in Table \ref{table:evaluation} and the qualitative  results in Figs.~\ref{fig:vimeo} and \ref{fig:flower} suggest that our performance is far beyond that of the baselines. 

Our quantitative evaluation involves four image quality metrics: NRMSE, PSNR, SSIM~\cite{wang2004image}, and LPIPS~\cite{zhang2018unreasonable}. Table \ref{table:evaluation} shows quantitative comparisons, and it is evident that our method outperforms the baselines by a large margin in most of the settings, including combinations of different scale factors, frame gaps, and parallax. In general, our method achieves approximately 10dB gain of PSNR upon the baselines. 

We provide qualitative comparisons in Figs. \ref{fig:vimeo} and \ref{fig:flower}
respectively on the Vimeo90k, Flower, LFVideo datasets under the challenging scale $8\times$, largest parallax and largest frame gap setting. Firstly, benefiting from the reference image, the example-based approaches show better results than the automatic colorization approaches. Moreover, among the example-based approaches, our method generates the most coherent images with less color bleeding effects, showing the clear advantage of our algorithm.

\begin{wrapfigure}{r}{0.45\textwidth}
\vspace{-12mm}
  \begin{center}
    \includegraphics[width=0.45\textwidth]{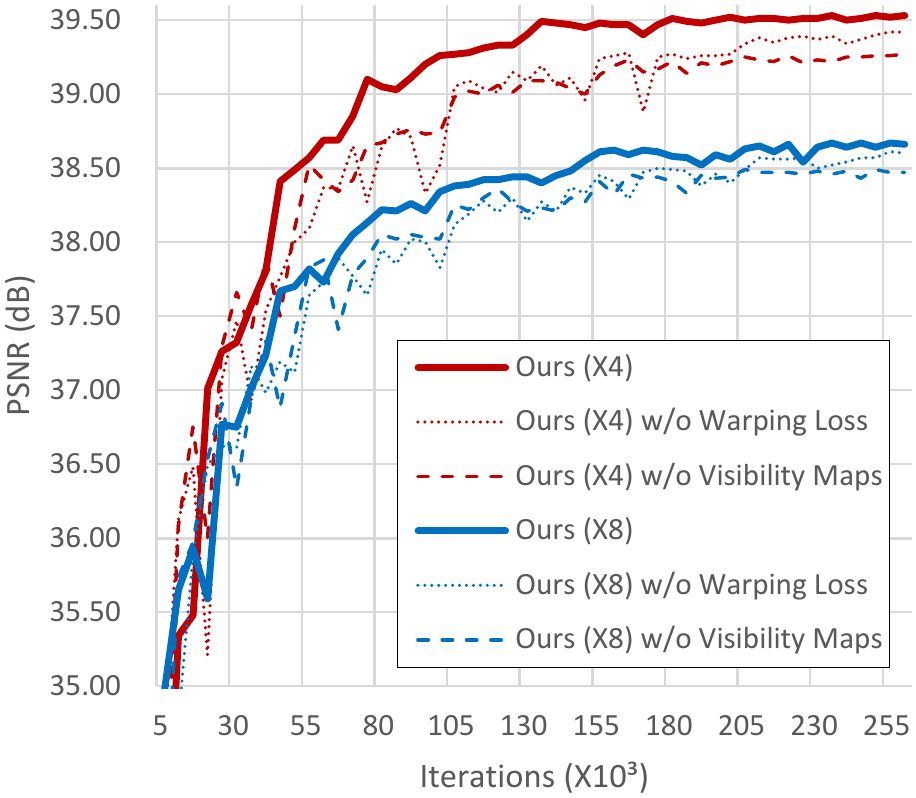}
  \end{center}
  \vspace{-3mm}
  \caption{The PSNR performance of the different variants with some of the components disabled during training on Vimeo90K dataset.}
\vspace{-12mm}
\label{fig:ablation}
\end{wrapfigure}


\vspace{-2mm}
\subsection{Ablation Study}
\vspace{-2mm}
This section investigates the role of proposed visibility maps and warping loss by two variants of our pipeline, which respectively turn off one of the components while remaining others. In particular, we run our test on the Vimeo90k dataset, with resolution gaps being $4\times$ and $8\times$.
According to Fig.~\ref{fig:ablation}, disabling either component leads to a performance drop, suggesting the necessities of both. 
\vspace{-2mm}
\paragraph{Visibility Maps} When visibility maps are disabled,
slight degradation in PSNR is observed in Fig.  \ref{fig:ablation}. Also, as shown in Fig.~\ref{fig:visibility}, artifacts are more evident without visibility maps, suggesting that visibility maps can reduce the wrong colorization due to inconsistent visibility between input images.

\vspace{-2mm}
\paragraph{Warping Loss} From Fig.~\ref{fig:ablation}, the performance of our network drops if warping loss is remove. We speculate that it is because the colorization loss is for image synthesis and does not explicitly define terms for flow estimation. In contrast, the warping loss is necessary to
regularize the flow estimator training and enable better convergence.

%% file: section/conclude.tex
\vspace{-2mm}
\section{Conclusion}
\label{sec:conlusion}
\vspace{-2mm}
We propose a novel convolutional neural network to facilitate the trade-off between spatial-temporal resolution and color depth in the imaging system. Our method fuses data to obtain an HR color image given cross-domain and cross-scale input pairs captured by the color-plus-mono dual camera. In contrast to previous works, our method can adapt to color and monochrome cameras with the various spatial-temporal resolution; thus it enables more flexible and generalized imaging systems. The key ingredient of our method is a cross-camera alignment module that generates multi-scale corre-spondence for cross-domain image alignment. 
Experiments on several datasets demonstrate the superior performance of our method (around 10dB in PSNR) compared to state-of-the-art methods. 